\pdfoutput=1

\documentclass[11pt]{article}

\usepackage[preprint]{acl}

\usepackage{times}
\usepackage{latexsym}
\usepackage{float}

\usepackage[T1]{fontenc}

\usepackage[utf8]{inputenc}

\usepackage{microtype}

\usepackage{inconsolata}

\usepackage{makecell}
\usepackage{amsmath}
\usepackage{amssymb}
\usepackage{booktabs}
\usepackage{pifont}
\usepackage{times}
\usepackage{soul}
\usepackage{url}
\usepackage{multirow}
\usepackage[utf8]{inputenc}
\usepackage{wrapfig}
\usepackage{tcolorbox}
\usepackage{listings}
\usepackage{framed}
\usepackage{textcomp}
\usepackage{minitoc}
\usepackage{colortbl}
\usepackage{xcolor}
\usepackage{algorithm}
\usepackage{algpseudocode}
\usepackage{amssymb}
\usepackage{xcolor}
\usepackage{pifont}
\usepackage{booktabs}  
\usepackage{colortbl}  
\usepackage{multirow}  
\usepackage{array}
\usepackage{graphicx}  
\usepackage{xcolor}    
\usepackage{pifont}    
\usepackage{tikz}
\usepackage{hyperref}
\usepackage{minitoc}
\usepackage{subcaption}

\usepackage[capitalize]{cleveref}
\crefname{section}{Sec.}{Secs.}
\Crefname{section}{Section}{Sections}
\Crefname{table}{Table}{Tables}
\crefname{table}{Tab.}{Tabs.}

\title{FedNano: Toward Lightweight Federated Tuning for Pretrained Multimodal Large Language Models}

\author{Yao Zhang \textsuperscript{\rm 1,4}\thanks{Equal contribution} \thanks{Corresponding authors}
\qquad Hewei Gao \textsuperscript{\rm 2}\footnotemark[1] 
\qquad Haokun Chen\textsuperscript{\rm 1,3} \\
\qquad \bf Weiguo Li \textsuperscript{\rm 5}
\qquad \bf Yunpu Ma \textsuperscript{\rm 1,4} \footnotemark[2]
\qquad \bf Volker Tresp\textsuperscript{\rm 1,4}\footnotemark[2]\\
    \textsuperscript{\rm 1} LMU Munich \qquad  
    \textsuperscript{\rm 2} Technical University of Munich \qquad  
    \textsuperscript{\rm 3} Siemens Technology \\
    \textsuperscript{\rm 4} Munich Center for Machine Learning \qquad  
    \textsuperscript{\rm 5} University Heidelberg \\
    yzhang@dbs.ifi.lmu.de \qquad yunpu.ma@ifi.lmu.de \qquad tresp@dbs.ifi.lmu.de
   	}

\begin{document}
\maketitle

\begin{abstract}
Multimodal Large Language Models (MLLMs) excel in tasks like multimodal reasoning and cross-modal retrieval but face deployment challenges in real-world scenarios due to distributed multimodal data and strict privacy requirements. Federated Learning (FL) offers a solution by enabling collaborative model training without centralizing data. However, realizing FL for MLLMs presents significant challenges, including high computational demands, limited client capacity, substantial communication costs, and heterogeneous client data. Existing FL methods assume client-side deployment of full models, an assumption that breaks down for large-scale MLLMs due to their massive size and communication demands. To address these limitations, we propose \textbf{\textit{\textit{FedNano}}}, the first FL framework that centralizes the LLM on the server while introducing \textit{NanoEdge}, a lightweight module for client-specific adaptation. \textit{NanoEdge} employs modality-specific encoders, connectors, and trainable \textit{NanoAdapters} with low-rank adaptation. This design eliminates the need to deploy LLM on clients, reducing client-side storage by \textbf{95\%}, and limiting communication overhead to only \textbf{0.01\%} of the model parameters. By transmitting only compact \textit{NanoAdapter} updates, \textit{FedNano} handles heterogeneous client data and resource constraints while preserving privacy. Experiments demonstrate that \textit{FedNano} outperforms prior FL baselines, bridging the gap between MLLM scale and FL feasibility, and enabling scalable, decentralized multimodal AI systems.
\end{abstract}

\section{Introduction}
\begin{figure}[!t]
  \centering
  \includegraphics[width=0.5\textwidth]{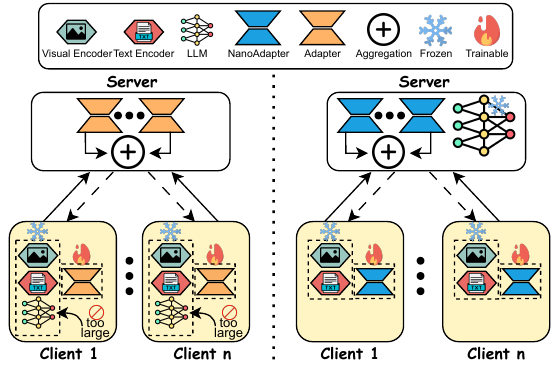}
  \caption{Comparison between traditional PEFT-based FL (left) and our proposed \textit{FedNano} (right). \textit{FedNano} keeps the LLM centralized on the server and performs lightweight tuning on clients, reducing both computation and communication overhead.}
  \label{intro}
  \vspace{-10pt}
\end{figure}

Multimodal Large Language Models (MLLMs) \cite{zhu2023minigpt, liu2024visual,peng2023kosmos,alayrac2022flamingo,li2023blip} excel in tasks like cross-modal retrieval \cite{yin2024survey}, making them indispensable for applications such as visual question answering (VQA) \cite{antol2015vqa}. However, real-world deployment remains fundamentally constrained: multimodal data is inherently decentralized and privacy-sensitive, while the large parameter footprint of MLLMs renders on-device execution infeasible for edge clients.

Federated learning (FL) \cite{mcmahan2017communication} offers a promising solution for decentralized multimodal training. However, applying FL to MLLMs presents fundamental system-level challenges. First, although parameter-efficient fine-tuning (PEFT) \cite{houlsby2019parameter, lester2021power, zaken2021bitfit, hu2021lora} reduces the number of trainable parameters, it still requires deploying the full MLLM—often exceeding 10 billion parameters—on each client, which is impractical for resource-constrained devices such as mobile phones or IoT systems. Second, PEFT methods typically insert adapters into internal layers of the language model, requiring structural access and full-model execution on clients, as seen in recent FL adaptations such as FedDPA-F \cite{yang2024dual}, pFedLoRA \cite{yi2023fedlora}, and FedIT \cite{zhang2024towards}. Third, the resulting adapter updates remain sizable, imposing substantial communication overhead across training rounds. Finally, non-IID client data introduces statistical heterogeneity that degrades global model convergence. These limitations collectively constrain the scalability and practicality of existing FL approaches for MLLMs.

To address these challenges, we propose \textbf{\textit{FedNano}}, the first FL framework that enables MLLM adaptation without deploying LLM on clients. As illustrated in Fig.~\ref{intro}, \textit{FedNano} centralizes LLM on the server in a frozen state, and equips each client with \textit{NanoEdge}—a lightweight adaptation module comprising modality-specific encoders, connectors, and trainable \textit{NanoAdapters}. These adapters operate externally to LLM and are optimized using low-rank decomposition \cite{hu2021lora}, minimizing both parameter size and transmission cost. This design removes the need for local LLM deployment, reduces client storage by over \textbf{95\%}, as shown in Tab. \ref{tab:num_params_comparison}. Only compact \textit{NanoAdapter} updates are exchanged across training rounds, achieving \textbf{over 99\% communication reduction} compared to PEFT-based FL methods \cite{chen2023feddatapproachfoundationmodel, yang2024dual}. By decoupling adaptation from the LLM, \textit{FedNano} provides a scalable and communication-efficient solution for real-world MLLM deployment.

To address client heterogeneity, \textit{FedNano} adapts Fisher Merging \cite{matena2022merging} to align global updates with client-specific data distributions. This adaptation improves performance on non-IID datasets and outperforms traditional aggregation methods such as FedAvg \cite{mcmahan2017communication} and FedProx \cite{li2020federated}. By integrating these innovations, \textit{FedNano} effectively bridges the gap between the computational complexity of MLLMs and the constraints of FL, enabling efficient deployment in real-world scenarios.

Experiments across diverse MLLM and multimodal tasks demonstrate that \textit{FedNano} not only outperforms existing methods but also significantly reduces resource and communication costs, enabling the scalable, efficient, and privacy-preserving deployment of MLLMs. This framework lays a strong foundation for advancing multimodal AI systems in decentralized real-world applications, including personalized healthcare, cross-device collaboration, and multimodal user interfaces.
	
\begin{table}[t]
\centering
\scalebox{0.90}{
\setlength{\tabcolsep}{4pt} 
\renewcommand{\arraystretch}{1.2}
\begin{tabular}{lcc}
\toprule
Approach       & Client Params          & Server Uploads \\ \midrule

\textit{FedNano}        & 304.55M (4.30\%)       & 1.05M (0.01\%)  \\ 
FedDPA-F       & 7222.81M (100\%)       & 180.89M (2.50\%)  \\ 
\textbf{Reduction} & \textbf{$\downarrow$ 95.7\%} & \textbf{$\downarrow$ 99.4\%} \\

\bottomrule
\end{tabular}}
\caption{Comparison of parameter distribution and communication efficiency between \textit{FedNano} and FedDPA-F \cite{yang2024dual} on LLaVA-1.5-7B \cite{liu2024visual}. \textit{Client Params} denotes parameters retained on client devices, while \textit{Server Uploads} denotes parameter updates sent to the server per round. Both use rank-64 adapters.}
\label{tab:num_params_comparison}
\vspace{-10pt}
\end{table}

The key contributions of this work are:\\
1. Novel FL Framework for MLLMs: We propose \textit{FedNano}, the first framework that centralizes the LLM on the server and enables lightweight client-side adaptation via \textit{NanoEdge}, reducing client storage by over \textbf{95\%} and enabling practical deployment on resource-constrained devices.\\
2.	Communication-Efficient Adaptation: \textit{FedNano} employs low-rank decomposition in \textit{NanoAdapters}, achieving an over \textbf{99\%} reduction in the number of transmitted parameters, allowing efficient deployment in bandwidth-constrained environments.\\
3. Improved Generalization on Non-IID Data: We adapt Fisher Merging for FL, aligning global updates with client-specific distributions to improve model performance on heterogeneous datasets.\\
4. Comprehensive Validation: Extensive experiments demonstrate the effectiveness and efficiency of \textit{FedNano}, establishing it as a scalable solution for real-world MLLM deployment.

\section{Related Work}
\subsection{Multimodal Large Language Models}
MLLMs \cite{zhu2023minigpt, liu2024visual, peng2023kosmos, alayrac2022flamingo, li2023blip, dai2023instructblipgeneralpurposevisionlanguagemodels} extend LLMs \cite{touvron2023llama, peng2023instruction, bai2023qwen} by integrating modality-specific encoders and connectors to process multimodal inputs. Recent works focus on efficient alignment, using lightweight connectors such as the linear projection in MiniGPT-4~\cite{zhu2023minigpt} or the MLP bridge in LLaVA~\cite{liu2024visual}. However, these models assume full model access, which is incompatible with federated settings due to privacy and resource constraints. \textit{FedNano} resolves this by freezing the LLM on the server and enabling lightweight client-side adaptation via \textit{NanoAdapters}.

\subsection{Parameter Efficient Fine-tuning}
PEFT techniques \cite{houlsby2019parameter, lester2021power, zaken2021bitfit, hu2021lora} adapt large pretrained models by updating only a small set of parameters, significantly reducing training costs. They include additive methods like adapters \cite{houlsby2019parameter} and soft prompts \cite{lester2021power}, selective tuning such as BitFit \cite{zaken2021bitfit}, and reparameterization methods like LoRA \cite{hu2021lora}. While effective in centralized settings, PEFT-based FL methods~\cite{chen2023feddatapproachfoundationmodel, wang2024flora, zhang2024towards} assume the full model, including LLM, can be deployed on clients. This becomes impractical for MLLMs, where LLM accounts for the vast majority of parameters and cannot be hosted on resource-limited devices. To overcome this, \textit{FedNano} introduces a new paradigm: the LLM is frozen and centralized on the server, while lightweight \textit{NanoAdapters} are deployed on clients. This design eliminates the need for full-model access, reduces client overhead, and enables scalable FL for MLLM. Unlike conventional PEFT, which are inserted into LLM, \textit{NanoAdapters} operate externally, interfacing solely through the modality connector. This allows adaptation without modifying or executing LLM on clients.

\subsection{Multimodal Federated Learning}
Multimodal FL has gained increasing attention for handling data heterogeneity and privacy constraints in real-world deployments. Prior work has focused on vision-language models, proposing strategies for modality imbalance \cite{yu2023multimodalfederatedlearningcontrastive, che2024leveragingfoundationmodelsmultimodal}, non-IID distributions \cite{yang2024dual, zhang2024towards, chen2024disentanglement}, and client personalization \cite{yi2023fedlora, chen2023feddatapproachfoundationmodel}. Benchmarks like FedMultimodal \cite{feng2023fedmultimodalbenchmarkmultimodalfederated} and FedMLLM \cite{xu2024fedmllm} further standardize evaluation in heterogeneous multimodal settings. However, these methods still rely on client-side full model deployment. For MLLMs, this becomes infeasible due to their scale. Even with PEFT, deploying full MLLMs locally remains out of reach, and transmitting adapter updates still incurs significant communication overhead. \textit{FedNano} departs from this design by keeping the LLM on the server and transmitting only compact \textit{NanoAdapter} updates from clients. This makes it the first scalable FL framework tailored for large-scale MLLMs, enabling efficient multimodal collaboration without sacrificing practicality.

\section{Methodology}
\begin{figure*}[t]
      \includegraphics[width=\textwidth]{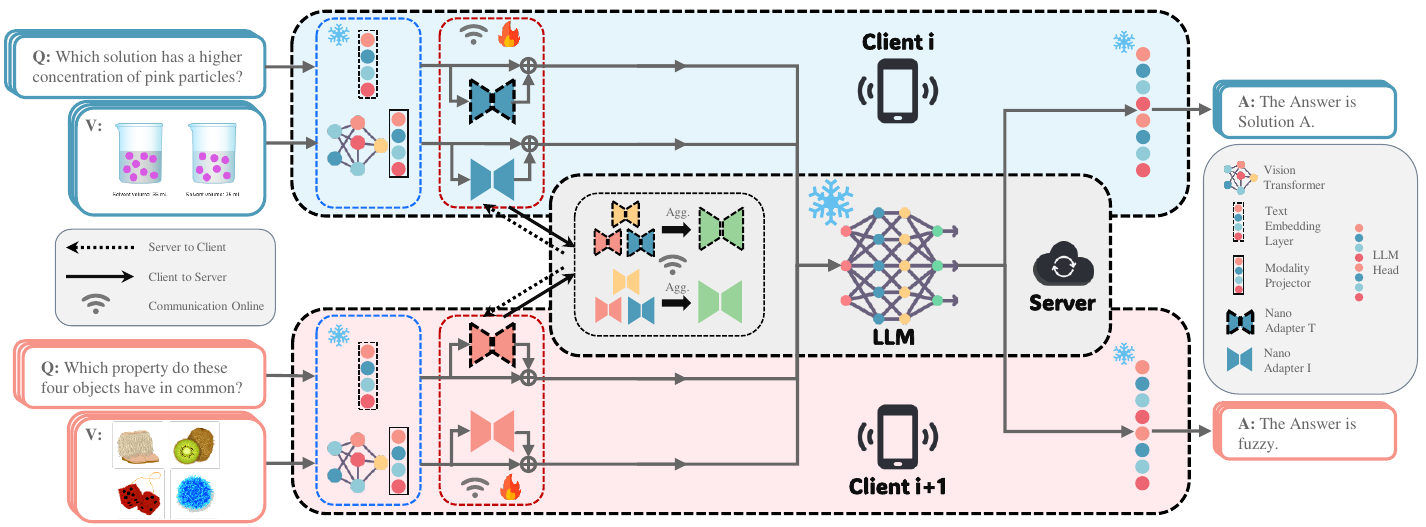}
      \caption{Overview of the \textit{FedNano} framework. The server hosts the frozen LLM, while each client performs local tuning via \textit{NanoEdge}, which includes \textit{NanoAdapter}-T for text and \textit{NanoAdapter}-I for vision. Clients upload low-rank adapter updates, which are aggregated on the server using Fisher merging. This design reduces client overhead and supports scalable, multimodal federated learning under data heterogeneity.}
      \label{fig:main_fig}
      \vspace{-5pt}
    \end{figure*}
    
\subsection{Problem Definition}
This work addresses federated fine-tuning for MLLMs in decentralized environments with statistical data heterogeneity. Each client $k$ holds a private multimodal dataset $D_k = \{(v_k^i, q_k^i, a_k^i)\}$, comprising image-question-answer triplets. We assume complete modality availability and a shared model architecture across all clients; only data distributions differ~\cite{chen2023feddatapproachfoundationmodel}. The marginal distributions of $v_k^i$, $q_k^i$, and $a_k^i$ vary across clients, resulting in shifts in both visual and textual representations, as well as answer semantics. Such heterogeneity poses challenges for achieving consistent generalization, as standard aggregation strategies struggle to align diverse local updates.

Our objective is to collaboratively fine-tune a shared global MLLM for VQA \cite{antol2015vqa}. Following \cite{liu2024improved}, we formulate this as an open-ended generation problem, where the model generates free-form answers given image-question pairs. Existing approaches assume that the full MLLM can be deployed on each client, which is infeasible in practice due to the massive size of LLM backbones. Client devices often lack sufficient compute, memory, and bandwidth to support such models, and privacy regulations further restrict centralized data access. These constraints call for a new FL framework that avoids client-side LLM deployment while enabling efficient adaptation and communication. To address these challenges, we propose \textbf{\textit{FedNano}}, a parameter-efficient framework that centralizes the computationally intensive LLM on the server while enabling lightweight, client-specific tuning. In the following sections, we detail the design of \textit{FedNano}, focusing on how it minimizes computational and communication overhead and addresses data heterogeneity.

\subsection{Overview of \textit{FedNano} Architecture} 

\textit{FedNano} is designed to address the inherent challenges of deploying MLLMs in FL environments. As shown in Fig.~\ref{fig:main_fig}, it introduces a new architecture that centralizes the computationally intensive LLM on the server, while clients maintain only lightweight \textit{NanoEdge} modules for task-specific adaptation. \textit{NanoEdge} freezes the modality encoders and connector, restricting training to task-specific \textit{NanoAdapters}. This design eliminates the need to deploy the full model on resource-constrained devices, reducing client-side computation and enabling edge deployment on mobile or IoT systems. The complete training and aggregation algorithm is provided in Appendix~\ref{app:code}.

\textit{FedNano} jointly addresses three key challenges in MLLM-based FL: high computation, communication cost, data heterogeneity. By offloading the LLM to the server, clients train only the \textit{NanoEdge} module, which includes frozen encoders and a connector, and optimizes a small set of \textit{NanoAdapters} for task-specific adaptation. The total client-side module accounts for less than \textbf{5\%} of the model parameters, while the trainable \textit{NanoAdapters} comprise only \textbf{0.01\%}. During aggregation, only \textit{NanoAdapters} updates are uploaded, significantly reducing communication overhead. \textit{NanoAdapters} are optimized via low-rank decomposition, enabling expressive local tuning while preserving pretrained alignment with the frozen LLM. This compact update mechanism supports low-bandwidth environments and enhances training efficiency. To address data heterogeneity, \textit{FedNano} integrates Fisher Merging \cite{matena2022merging} into FL as an advanced aggregation strategy, leveraging client-specific posterior estimates to align local updates with global objectives. By weighting and combining \textit{NanoAdapter} updates based on their estimated importance, this method improves robustness across diverse datasets, even under non-IID conditions. Together with its architectural and optimization designs, \textit{FedNano} bridges the gap between the computational barriers of MLLM deployment and the practical constraints of FL, offering a scalable, efficient, and privacy-preserving solution for decentralized multimodal learning.

\subsection{\textit{NanoEdge}: Client-Side Tuning Module}
MLLMs are composed of three key components: modality encoders, a connector, and a pretrained LLM backbone. The modality encoders extract embeddings from raw inputs, such as images and text, while the connector aligns these embeddings into a unified representation compatible with the LLM. Together, these components enable MLLMs to effectively handle diverse multimodal tasks by leveraging their pretrained capabilities.

Building on this structure, \textit{NanoEdge} introduces \textit{NanoAdapters} at the interface between the connector and the LLM to facilitate efficient task-specific tuning while preserving the pretrained alignment across modalities. By freezing the modality encoders and the connector, \textit{NanoEdge} maintains their alignment with the LLM, ensuring the foundational structure of the pretrained model remains intact. This design allows \textit{NanoAdapters} to focus solely on learning task-specific patterns from local client data and integrating federated knowledge updates, avoiding any disruption to the pretrained alignment. By restricting training to the lightweight \textit{NanoAdapter} parameters, \textit{NanoEdge} minimizes client-side computational demands while enabling efficient and privacy-preserving adaptation.

The \textit{NanoAdapters} employ a low-rank decomposition mechanism, inspired by LoRA \cite{hu2021lora}, consisting of a down-projection to reduce embedding dimensionality and an up-projection to restore it. This design balances parameter efficiency and adaptation capability, enabling \textit{NanoEdge} to perform localized tuning and transmit updates efficiently. Each modality is equipped with a dedicated \textit{NanoAdapter}—$\mathcal{A}_I$ for images and $\mathcal{A}_T$ for text—capturing modality-specific patterns essential for multimodal tasks. Unlike traditional adapters that are inserted into LLM, \textit{NanoAdapters} remain externally attached to the modality connector, requiring no structural access to or execution of LLM. This makes them uniquely compatible with server-hosted LLMs in federated environments.

\setlength{\tabcolsep}{5pt}
\begin{table*}[!t]
\centering
\scalebox{0.83}{
\begin{tabular}{llcccccccccccc}
\toprule
\multirow{2}{*}{Backbone} & \multirow{2}{*}{Approach} & \multicolumn{6}{c}{ScienceQA (Clients)} & \multicolumn{6}{c}{IconQA (Clients)} \\
\cmidrule(lr){3-8} \cmidrule(lr){9-14}
 &  & C1 & C2 & C3 & C4 & C5 & Avg & C1 & C2 & C3 & C4 & C5 & Avg \\
\midrule
\multirow{6}{*}{MiniGPT-4} 
    & Centralized     & 73.70 & 88.34 & 89.83 & 84.52 & 87.41 & 84.76 & 80.76 & 86.62 & 81.16 & 82.74 & 85.36 & 83.33 \\
    & LocFT           & 67.74 & 74.69 & 77.42 & 72.46 & 74.07 & 73.28 & 67.70 & 73.48 & 70.63 & 70.86 & 77.53 & 72.04 \\
    & FedAvg         & 70.22 & 79.65 & 79.65 & 75.19 & 75.56 & 76.05 & 70.31 & 75.61 & 74.98 & 72.76 & 81.25 & 74.98 \\
    & FedProx        & 70.97 & 80.40 & 80.15 & 75.19 & 75.80 & 76.50 & 70.94 & 77.36 & 74.58 & 71.50 & 80.70 & 75.01 \\
    & FedDPA-F      & \textbf{71.96} & 78.41 & \textbf{81.14} & 76.42 & 75.80 & 76.75 & 70.94 & \textbf{77.91} & 74.51 & 73.08 & 80.30 & 75.35 \\
    & \textbf{\textit{FedNano}}& 68.98 & \textbf{81.89} & 80.89 & \textbf{76.43} & \textbf{77.04} & \textbf{77.05} & \textbf{72.21} & 77.28 & \textbf{75.85} & \textbf{74.27} & \textbf{82.52} & \textbf{76.42} \\
\midrule
\multirow{6}{*}{LLaVA-1.5} 
    & Centralized     & 83.87 & 91.07 & 89.33 & 90.57 & 89.38 & 88.84 & 86.62 & 88.92 & 84.88 & 87.25 & 88.45 & 87.22 \\
    & LocFT           & 71.96 & 80.89 & 76.92 & 79.65 & 75.80 & 77.04 & \textbf{75.93} & 78.94 & 72.53 & 74.35 & 76.50 & 75.65 \\
    & FedAvg        & 73.20 & \textbf{84.37} & 83.62 & 82.13 & \textbf{80.49} & 80.76 & 71.18 & 79.89 & 76.80 & \textbf{77.51} & \textbf{83.23} & 77.72 \\
    & FedProx       & 73.95 & \textbf{84.37} & 83.87 & 81.39 & 80.00 & 80.71 & 70.23 & 80.13 & 76.72 & \textbf{77.51} & 82.36 & 77.39 \\ 
    & FedDPA-F       & 73.70 & 84.12 & 84.12 & 81.89 & 79.51 & 80.67 & 72.12 & 79.65 & 76.80 & 77.43 & 82.36 & 77.68 \\
    & \textbf{\textit{FedNano}}& \textbf{74.94} & 84.12 & \textbf{84.86} & \textbf{82.88} & 80.25 & \textbf{81.41} & 72.13 & \textbf{80.44} & \textbf{77.36} & 77.43 & 82.83 & \textbf{78.04} \\
\bottomrule
\end{tabular}}
\caption{Performance comparison. Results include centralized training, local fine-tuning (LocFT), and various federated approaches. \textit{FedNano} achieves superior average performance on both datasets compared to other federated approaches, demonstrating its effectiveness in handling client heterogeneity.}
\label{tab:combined-results}
\vspace{-5pt}
\end{table*}

\subsection{Fisher-Guided Adaptive Aggregation}
In FL, model aggregation can be interpreted as maximizing the joint posterior likelihood across clients. Traditional methods like FedAvg implicitly assume isotropic Gaussian posteriors \cite{matena2022merging}, which oversimplifies client uncertainty and leads to degraded performance under data heterogeneity. \textit{FedNano} addresses this limitation by adopting Fisher Merging \cite{matena2022merging}, which leverages the Laplace approximation for more accurate posterior estimation. The global update is computed as:
\begin{equation}
\label{eq:FIM}
\theta_{global} = \frac{\sum_{k=1}^K \frac{|D_k|}{\sum_{k=1}^K|D_k|} F_k \theta_k}{\sum_{k=1}^K \frac{|D_k|}{\sum_{k=1}^K|D_k|} F_k},
\end{equation}
where $\theta_k$ denotes the \textit{NanoAdapter} parameters of client $k$, $F_k$ is the Fisher Information Matrix (FIM), which serves as the precision matrix of the Laplace approximation, and $D_k$ is the local dataset. This weighting improves the alignment of local updates with their estimated importance, enhancing generalization under non-IID data. To ensure scalability, \textit{FedNano} approximates the full FIM with its diagonal \cite{kirkpatrick2017overcoming}, and computes it efficiently from squared gradients during backpropagation \cite{wu2023pi}. This reduces computation from $O(|\theta|^2)$ to $O(|\theta|)$ without sacrificing aggregation accuracy. Compared to uniform averaging, this method dynamically prioritizes impactful updates, achieving stronger global performance under statistical heterogeneity.

\section{Experiment}

\subsection{Experimental Setup}
We evaluate our approach on the Visual Question Answering (VQA) task using two established benchmarks: ScienceQA \cite{lu2022learn} and IconQA \cite{lu2021iconqa}. These datasets were selected for their well-defined categorical structures and multimodal complexities, making them particularly suitable for assessing the performance of FL in non-IID settings. To simulate FL in a non-IID setting, we partitioned the datasets using Dirichlet distributions following \cite{che2023federated, lai2022fedscale, zhang2024towards} with a concentration parameter $\alpha=1$ to create strongly non-IID splits. Partitioning was guided by topic annotations in ScienceQA and skill annotations in IconQA, ensuring heterogeneous yet meaningful distributions across five simulated clients. Each partition, representing an individual client dataset, maintains consistent train-validation-test splits for evaluation. We evaluate our approach on MiniGPT-4 \cite{zhu2023minigpt} and LLaVA-1.5 \cite{liu2024visual}. 

\subsection{Implementation Details}

\paragraph{Baselines} 
To the best of our knowledge, \textit{FedNano} is the first FL framework specifically designed to support MLLMs by centralizing the LLM on the server. This architectural shift renders existing PEFT-based FL methods inapplicable, as they assume full-model access and local integration with the LLM. Given the absence of prior work addressing this setting, we evaluate \textit{FedNano} against three representative FL baselines: FedAvg \cite{mcmahan2017communication}, a foundational aggregation method with limited handling of data heterogeneity; FedProx \cite{li2020federated}, which mitigates client drift through a proximal term but lacks parameter-specific adaptation; and FedDPA-F \cite{yang2024dual}, which integrates advanced alignment strategies but incurs high computational and communication overheads. We further include comparisons with a centralized model, representing the performance upper bound achieved with access to all data, and locally fine-tuned models, which operate in isolation without collaboration. 

\paragraph{Training Configurations} 
The training process includes 10 communication rounds ($R=10$), with each client performing one local epoch per round using a batch size of 8. All experiments were conducted on NVIDIA A100 80G GPUs. 

\begin{table*}[!t]
\centering
\scalebox{0.94}{
\begin{tabular}{lcccccc|cccccc}
\toprule
\multirow{2}{*}{Approach} & \multicolumn{6}{c|}{ $\alpha=0.1$} & \multicolumn{6}{c}{$\alpha=5$} \\
\cmidrule(lr){2-7} \cmidrule(lr){8-13}
& C1 & C2 & C3 & C4 & C5 & Avg & C1 & C2 & C3 & C4 & C5 & Avg \\
\midrule
LocFT           & 69.94 & 75.80 & 75.48 & 73.18 & 77.00 & 74.28 & 65.71 & 70.62 & 71.41 & 72.76 & 70.64 & 70.22 \\
FedAvg          & 72.80 & 76.80 & 75.50 & 73.20 & 73.60 & 74.38 & 74.34 & 75.61 & 72.92 & \textbf{76.08} & 74.68 & 74.72 \\
FedProx         & 71.54 & 74.79 & 74.15 & 69.72 & 70.06 & 73.05 & 68.48 & 70.30 & 70.15 & 70.15 & 71.04 & 70.02 \\
FedDPA-F        & 70.25 & 76.40 & 74.10 & 72.50 & \textbf{78.55} & 74.27 & 71.52 & \textbf{76.83} & \textbf{74.51} & 73.24 & \textbf{75.84} & 74.38 \\
\textbf{\textit{FedNano}}         & \textbf{73.85} & \textbf{78.22} & \textbf{80.14} & \textbf{76.28} & 74.94 & \textbf{76.68} & \textbf{74.90} & 76.16 & 74.18 & 74.82 & 73.73 & \textbf{74.75} \\
\bottomrule
\end{tabular}}
\caption{Performance of MiniGPT-4 on IconQA under different data heterogeneity levels. \textit{FedNano} consistently outperforms all baselines, with the largest gains under highly non-IID conditions.}
\label{tab:diffd}
\vspace{-5pt}
\end{table*}

\subsection{Main Results}
Results in Tab.~\ref{tab:combined-results} demonstrate that FL methods consistently outperform locally fine-tuned models (LocFT), emphasizing the benefit of global knowledge sharing in distributed, heterogeneous settings.

\textit{FedNano} achieves the highest average performance among all FL methods, more effectively narrowing the gap to centralized training than existing baselines. While FedAvg performs competitively with simple weighted averaging, its inability to adapt to non-IID data results in suboptimal performance under heterogeneous distributions. FedProx mitigates client drift by constraining local updates toward the global model, but this rigid constraint limits flexibility, making it insufficient for complex multimodal tasks. FedDPA-F, though designed for personalization, requires careful tuning of global training epochs and risks overwriting the global adapter during local updates, potentially degrading performance due to catastrophic forgetting.

In contrast, the superior performance of \textit{FedNano} is attributed to its novel design and optimization strategies. As shown in Tab. \ref{tab:combined-results}, \textit{FedNano} achieves an average accuracy of 77.05\% on ScienceQA and 76.42\% on IconQA for MiniGPT-4, exceeding FedAvg and FedProx, indicating improved generalization in heterogeneous client environments. For LLaVA, \textit{FedNano} attains 81.41\% on ScienceQA and 78.04\% on IconQA, surpassing FedDPA-F and FedProx, demonstrating enhanced robustness in multimodal FL. These results validate the effectiveness of \textit{NanoAdapters} for modality-specific adaptation, while substantially reducing client-side computational and storage demands, enabling deployment on resource-limited devices. Moreover, \textit{FedNano} integrates Fisher Merging with a diagonal approximation of the FIM, allowing the system to prioritize critical parameter updates based on client-specific confidence. This results in more effective aggregation than uniform averaging, improving stability under non-IID distributions while reducing overfitting to local client noise. By balancing generalization and personalization, \textit{FedNano} consistently delivers strong performance across diverse client settings, all while maintaining minimal communication overhead.

\subsection{Analysis}

\begin{table*}[!t]
\centering
\scalebox{0.9}{
\begin{tabular}{lcccccccccccc}
\toprule
Approach & C1 & C2 & C3  & C4 & C5 & C6 & C7 & C8 & C9 & C10 & Avg \\
\midrule
LocFT       & 67.56 & 69.77 & 73.89 & 67.24 & 79.90 & 72.15 & 69.77 & 64.71 & 71.67 & 67.35 & 70.40 \\
FedAvg  & 74.52 & 81.01 & 78.00 & 78.63 & 85.91 & 79.90 & 75.94 & 75.63 & 70.90 & 77.86 & 77.83 \\
FedProx      & 73.89 & 76.74 & 77.37 & 75.63 & 84.01 & 76.58 & 73.41 & 71.36 & 78.79 & 72.29 & 76.00 \\
FedDPA-F        & 74.52 & 81.01 & 78.00 & 78.63 & 85.91 & 79.90 & 75.94 & 75.63 & 70.90 & 77.86 & 77.83 \\
\textbf{\textit{FedNano}}              & \textbf{77.03} & \textbf{82.77} & \textbf{78.22} & \textbf{79.67} & \textbf{88.57} & \textbf{80.35} & \textbf{81.34} & \textbf{72.84} & \textbf{73.77} & \textbf{79.47} & \textbf{78.86} \\
\bottomrule
\end{tabular}}
\caption{Performance of MiniGPT-4 on IconQA with 10 simulated clients. \textit{FedNano} achieves the best average accuracy, demonstrating strong scalability to larger federated setups.}
\label{tab:10client}
\vspace{-5pt}
\end{table*}

\paragraph{Robustness under Data Heterogeneity}
To assess the robustness of \textit{FedNano} under varying levels of data heterogeneity, we evaluate its performance on IconQA using the MiniGPT-4 backbone across different Dirichlet concentration values ($\alpha=0.1$ and $\alpha=5$). As shown in Tab.~\ref{tab:diffd}, \textit{FedNano} consistently achieves the highest average accuracy in the highly non-IID setting ($\alpha=0.1$), outperforming all FL baselines. This demonstrates the effectiveness of its Fisher-guided aggregation in aligning heterogeneous client updates. While the performance gap narrows under near-IID conditions ($\alpha=5$), \textit{FedNano} remains competitive, indicating that its advantages are most pronounced in realistic heterogeneous federated scenarios.
 
\paragraph{Scalability to Larger Client Populations}
To evaluate the scalability of \textit{FedNano}, we extend the number of clients from 5 to 10 on the IconQA dataset using the MiniGPT-4 backbone. As shown in Tab.~\ref{tab:10client}, \textit{FedNano} achieves the highest average accuracy, consistently outperforming all baselines. This demonstrates that the framework retains its effectiveness even as the federated environment becomes more fragmented. The results confirm that \textit{FedNano} scales robustly with increasing client population, reinforcing its practicality for real-world large-scale federated deployments.

\paragraph{Generalization under Cross-Task Client Distribution}
We evaluate \textit{FedNano} in a challenging cross-task setup where four clients are respectively assigned A-OKVQA, OK-VQA, IconQA, and GQA, introducing significant task-level heterogeneity. As shown in Tab.~\ref{tab:cross-task}, \textit{FedNano} achieves stable and strong performance across all clients. This robustness stems from its modular design and Fisher-guided aggregation, which enable effective alignment of heterogeneous updates and support generalization across semantically diverse tasks.
\begin{table}[!t]
\centering
\scalebox{0.9}{
\begin{tabular}{lccccc}
\toprule
Approach & C1 & C2 & C3 & C4 & Avg \\
\midrule
FedAvg       & 34.35 & 28.83 & 29.00 & 29.53 & 30.86 \\
FedProx      & 52.45 & 50.82 & 59.80 & 42.15 & 51.30 \\
FedDPA-F     & 52.76 & 51.12 & 60.10 & 42.46 & 51.61 \\
\textbf{\textit{FedNano}} & \textbf{54.20} & \textbf{52.60} & \textbf{60.36} & \textbf{43.32} & \textbf{52.62} \\
\bottomrule
\end{tabular}}
\caption{Performance under cross-task federated setup on MiniGPT-4. \textit{FedNano} achieves the best average accuracy across clients with distinct VQA tasks.}
\label{tab:cross-task}
\vspace{-5pt}
\end{table}

\paragraph{The Necessity of Combining Both $\mathcal{A}_T$ and $\mathcal{A}_I$}

\begin{table}[!t]
\centering
\scalebox{0.9}{
\setlength{\tabcolsep}{4pt}
\renewcommand{\arraystretch}{1.2}
\begin{tabular}{lccc}
\toprule
Backbone       & Variants           & ScienceQA    & IconQA    \\ \midrule
\multirow{3}{*}{MiniGPT-4} 
 & $\mathcal{A}_T$         & 45.91  & 57.77  \\ 
 & $\mathcal{A}_I$         & 74.57  & 75.17  \\ 
 & $\mathcal{A}_T$ + $\mathcal{A}_I$      & \textbf{76.42}  & \textbf{76.04}  \\ \midrule

\multirow{3}{*}{LLaVA-1.5} 
 & $\mathcal{A}_T$         & 50.08  & 48.15  \\ 
 & $\mathcal{A}_I$         & 77.03  & 77.12  \\ 
 & $\mathcal{A}_T$ + $\mathcal{A}_I$     & \textbf{78.04}  & \textbf{77.83}  \\ 
\bottomrule
\end{tabular}}
\caption{Performance of different adapters. Combining $\mathcal{A}_T$ and $\mathcal{A}_I$ consistently yields the best results across backbones, confirming their complementarity.}
\vspace{-5pt}
\label{tab:I or T ablation}
\end{table}
To evaluate the necessity of the textual adapter $\mathcal{A}_T$ and the visual adapter $\mathcal{A}_I$, we conduct ablation experiments using three configurations: $\mathcal{A}_T$ only, $\mathcal{A}_I$ only, and both. For MiniGPT-4, $\mathcal{A}_T$ achieves 45.91\% on ScienceQA and 57.77\% on IconQA, while $\mathcal{A}_I$ improves to 74.57\% and 75.17\%. Their combination further boosts accuracy to 76.42\% and 76.04\%, outperforming $\mathcal{A}_I$ alone by +1.85\% and +0.87\%. As shown in Tab. \ref{tab:I or T ablation}, similar trends are observed with LLaVA-1.5, confirming the robustness of combining both adapters. The poor performance of $\mathcal{A}_T$ alone suggests that textual inputs provide insufficient task-relevant information in these vision-centric VQA tasks. These results validate the dual-adapter design of \textit{NanoEdge}, where $\mathcal{A}_I$ handles visual adaptation and $\mathcal{A}_T$ enhances generalization.

\begin{table}[!t]
\centering
\scalebox{0.85}{
\setlength{\tabcolsep}{3pt}
\renewcommand{\arraystretch}{1.2}
\begin{tabular}{lccc}
\toprule
 Dataset & Variants & MiniGPT-4 & LLaVA-1.5 \\ \midrule
 \multirow{5}{*}{ScienceQA}
 & \textit{FedNano}     & \textbf{77.05} & \textbf{81.41} \\
 & \textit{FedNano-EF}  & 76.55          & 80.81          \\
 & FedAvg      & 76.05          & 80.76          \\
 & FedProx     & 76.50          & 80.71          \\
 & FedDPA-F    & 76.75          & 80.67          \\\midrule
 \multirow{5}{*}{IconQA}
 & \textit{FedNano}        & \textbf{76.42} & \textbf{78.04} \\
 & \textit{FedNano-EF}     & 76.04 & 77.83 \\ 
 & FedAvg         & 74.98 & 77.72 \\ 
 & FedProx        & 75.01 & 77.39 \\ 
 & FedDPA-F       & 75.35 & 77.68 \\ 
\bottomrule
\end{tabular}}
\caption{Performance comparison of \textit{FedNano} and \textit{FedNano-EF} on ScienceQA and IconQA. \textit{FedNano} achieves the highest accuracy, while \textit{FedNano-EF} offers a trade-off with reduced computational overhead, demonstrating strong performance across both datasets.}
\label{tab:fedfm_vs_fmef_transposed}
\vspace{-10pt}
\end{table}

\paragraph{Trade-offs in Fisher-Guided Adaptive Aggregation}
FIM is specific to a particular set of model parameters and plays a key role in the ability of \textit{FedNano} to achieve superior global alignment by capturing parameter importance. To compute the FIM precisely, \textit{FedNano} employs additional forward and backward passes per communication round, ensuring accurate parameter estimation. While this enhances accuracy, it introduces modest computational overhead. To explore the trade-offs between precision and efficiency, we conduct an ablation study with \textit{FedNano-EF}, a variant that approximates the FIM during standard training, eliminating the need for additional computation steps. This modification reduces computational overhead to the level of FedAvg. Despite this simplification, \textit{FedNano-EF} incurs only a slight accuracy trade-off and consistently outperforms baselines, as shown in Tab. \ref{tab:fedfm_vs_fmef_transposed}. These results demonstrate the adaptability of \textit{FedNano}: the standard version excels in accuracy-critical tasks by leveraging precise FIM computation to optimize alignment, while \textit{FedNano-EF} provides a practical alternative for resource-constrained environments, achieving strong performance with reduced overhead.

\begin{figure}[t]
    \centering
    \begin{subfigure}[b]{0.235\textwidth}
        \centering
        \includegraphics[width=\textwidth]{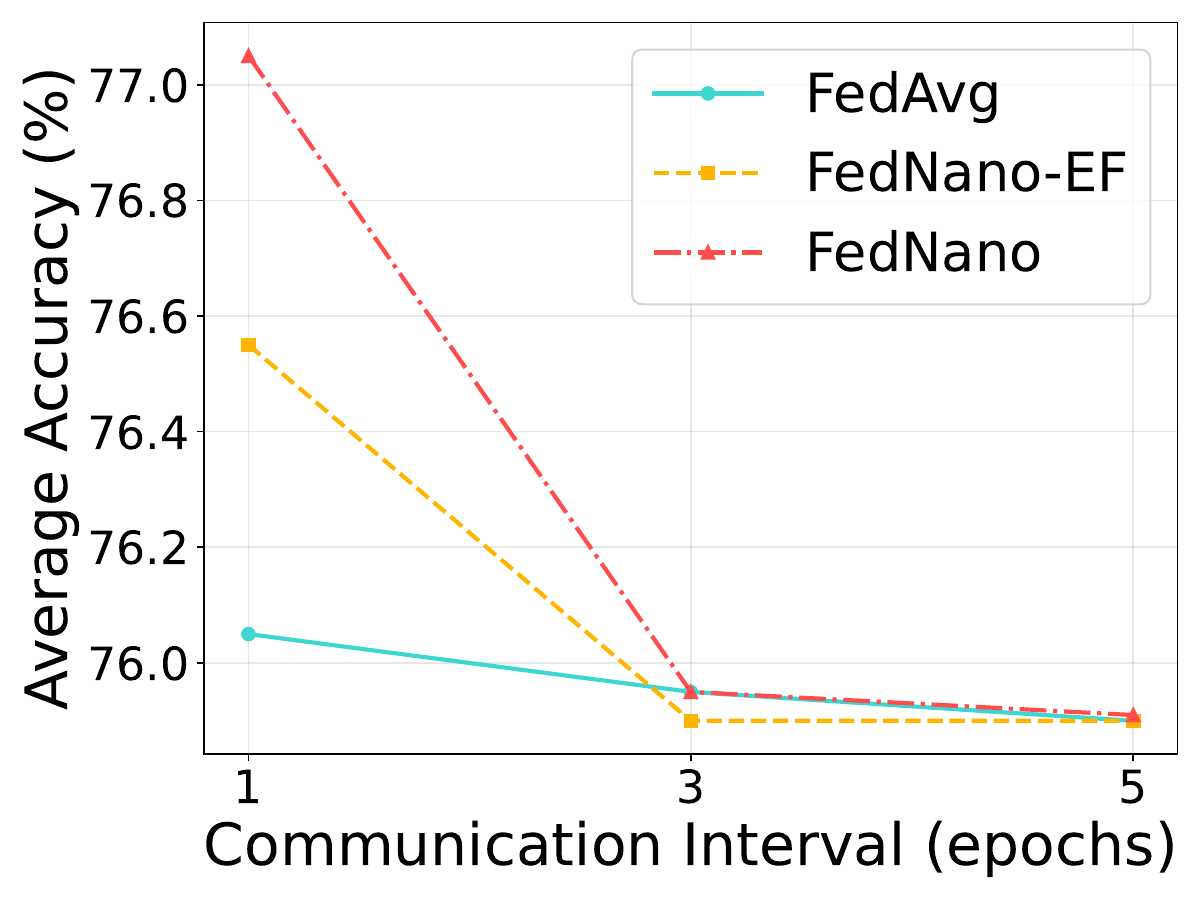}
        \caption{Communication Freq.}
        \label{fig:freq ablation experiments of MiniGPT-4 on ScienceQA}
    \end{subfigure}
    \hfill
    \begin{subfigure}[b]{0.235\textwidth}
        \centering
        \includegraphics[width=\textwidth]{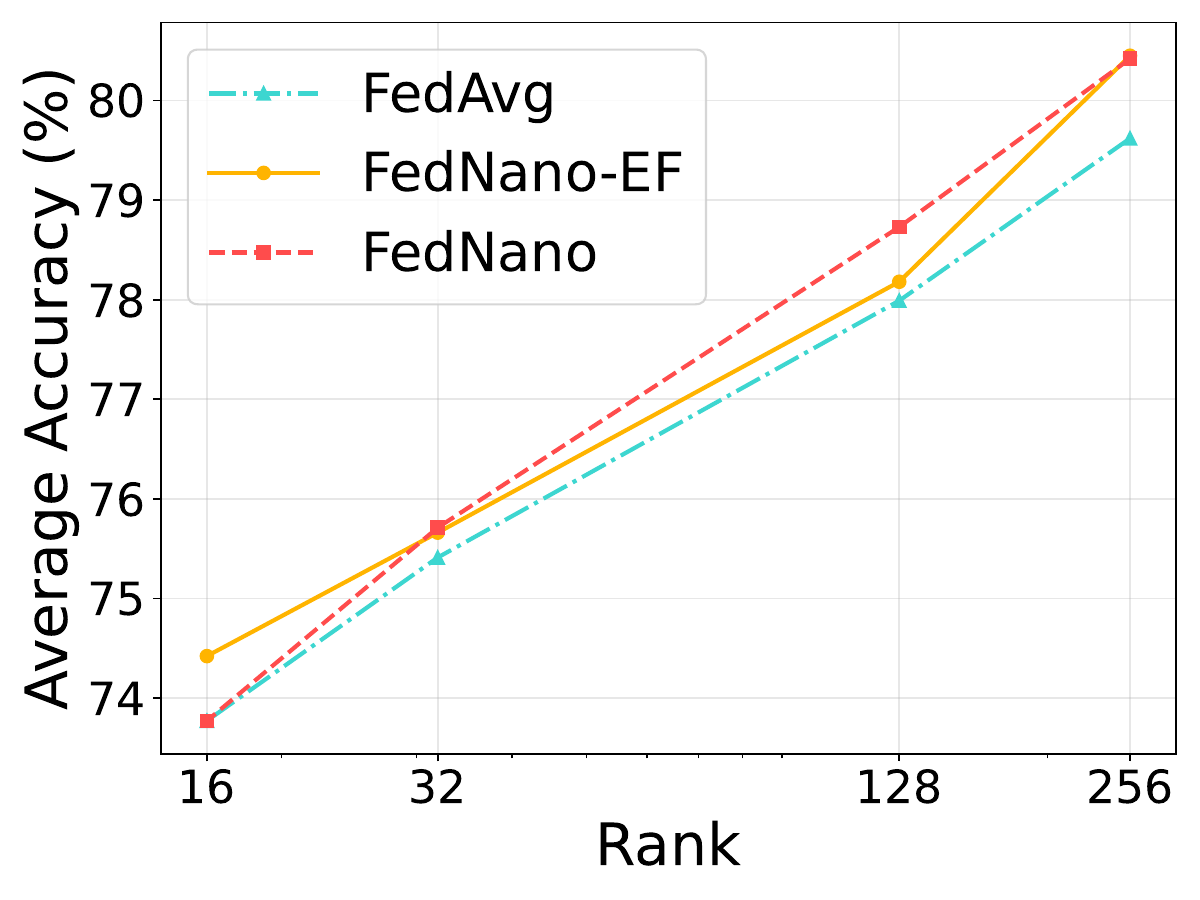}
        \caption{Adapter Rank}
		\label{fig:rank ablation experiments of MiniGPT-4 on ScienceQA}
    \end{subfigure}
    \caption{(a) Impact of communication frequency. \textit{FedNano} outperforms FedAvg, with more frequent communication amplifying its advantages; (b) Effect of adapter rank. \textit{FedNano} consistently achieves superior performance, demonstrating its ability to capture task-specific and client-specific information effectively.}
    \label{fig:combined_results}
    \vspace{-10pt}
\end{figure}

\paragraph{Frequent Communication Amplifies the Advantages of \textit{FedNano}}
As shown in Fig.~\ref{fig:freq ablation experiments of MiniGPT-4 on ScienceQA}, reduced communication frequency leads to a general decline in global model performance across all methods due to increased parameter divergence, which hinders effective aggregation. Importantly, the results highlight that \textit{FedNano} outperforms FedAvg by a larger margin when communication is more frequent. With shorter intervals, FIM mechanism of \textit{FedNano} can better leverage aligned client parameters to prioritize impactful updates, amplifying its advantages in handling data heterogeneity. In contrast, FedAvg struggles with parameter divergence regardless of communication frequency, showing minimal improvement with more frequent updates. These findings underscore that while frequent communication benefits all methods, it significantly enhances the effectiveness of \textit{FedNano}, reinforcing its superior ability to integrate client-specific updates and maintain robust performance in federated learning environments.

\paragraph{Higher Adapter Ranks Enhance \textit{FedNano} Performance}

Fig.~\ref{fig:rank ablation experiments of MiniGPT-4 on ScienceQA} illustrates the impact of adapter rank, comparing \textit{FedNano} with FedAvg on the ScienceQA dataset. As the adapter rank increases, accuracy improves due to the enhanced capacity to encode task-specific and client-specific information, which is particularly important in non-IID settings. However, higher ranks also incur greater communication costs, necessitating a trade-off between performance and resource efficiency in FL. \textit{FedNano} consistently outperforms FedAvg across all ranks, with the performance gap widening at higher ranks. This improvement is driven by the FIM aggregation, which leverages richer client-specific updates at higher ranks to achieve better alignment between local contributions and the global model. In contrast, at lower ranks, the limited adapter capacity constrains the quality of updates, reducing the effectiveness of FIM aggregation.

\section{Conclusion}
This work introduced \textit{FedNano}, an FL framework that tackles the unique challenges of deploying MLLMs in decentralized settings. By centralizing the LLM on the server and employing lightweight \textit{NanoAdapters} on clients, \textit{FedNano} achieves significant gains in both resource and communication efficiency, while effectively addressing data heterogeneity in non-IID environments. Comprehensive evaluations on ScienceQA and IconQA benchmarks demonstrate that \textit{FedNano} consistently outperforms state-of-the-art FL baselines, further narrowing the gap between federated and centralized training. By combining scalable design with robust performance, \textit{FedNano} offers a practical and privacy-preserving solution, advancing the real-world deployment of MLLMs.

\section*{Limitation and Future Work}
While \textit{FedNano} demonstrates strong performance and efficiency, several aspects remain open for further improvement. One limitation lies in the assumption that all clients possess similar hardware capabilities to manage \textit{NanoAdapters}, a condition that may not hold in real-world federated settings with highly heterogeneous devices. Future research could explore adaptive mechanisms that dynamically adjust \textit{NanoAdapter} configurations to fit each client’s resource constraints, enabling broader applicability across diverse edge platforms.

Although \textit{FedNano} is designed to address high heterogeneity across clients, practical deployments may involve incomplete modality settings, where some clients lack access to specific input modalities. Handling such partial data scenarios requires new strategies to ensure effective collaboration and generalization without assuming full modality availability. Moreover, while the current framework supports vision and language inputs, extending it to incorporate additional modalities such as audio, sensor data, or time-series streams could enable applications in multimodal healthcare, industrial IoT, and autonomous systems.

Another promising direction is integrating \textit{FedNano} into federated multi-agent systems, where distinct agents learn collaboratively, which could enable novel applications in domains like logistics and autonomous driving, further demonstrating the flexibility of the framework.

Finally, while \textit{FedNano} already offers strong privacy guarantees by transmitting only lightweight adapter updates, incorporating advanced privacy-preserving techniques such as differential privacy could provide even stronger safeguards. A key challenge will be achieving such privacy enhancements without sacrificing the computational and communication efficiency that underpins \textit{FedNano’s} practicality.

In summary, while \textit{FedNano} addresses key challenges in federated learning for MLLMs, these future directions highlight its potential for continued innovation. Enhancing its adaptability to incomplete modalities, extending cross-modal coverage, and incorporating stronger privacy mechanisms could further establish \textit{FedNano} as a foundational framework for real-world federated multimodal AI systems.

\bibliography{custom}

\begin{thebibliography}{35}
\providecommand{\natexlab}[1]{#1}

\bibitem[{Alayrac et~al.(2022)Alayrac, Donahue, Luc, Miech, Barr, Hasson, Lenc,
  Mensch, Millican, Reynolds et~al.}]{alayrac2022flamingo}
Jean-Baptiste Alayrac, Jeff Donahue, Pauline Luc, Antoine Miech, Iain Barr,
  Yana Hasson, Karel Lenc, Arthur Mensch, Katherine Millican, Malcolm Reynolds,
  et~al. 2022.
\newblock Flamingo: a visual language model for few-shot learning.
\newblock \emph{Advances in neural information processing systems},
  35:23716--23736.

\bibitem[{Antol et~al.(2015)Antol, Agrawal, Lu, Mitchell, Batra, Zitnick, and
  Parikh}]{antol2015vqa}
Stanislaw Antol, Aishwarya Agrawal, Jiasen Lu, Margaret Mitchell, Dhruv Batra,
  C~Lawrence Zitnick, and Devi Parikh. 2015.
\newblock Vqa: Visual question answering.
\newblock In \emph{Proceedings of the IEEE international conference on computer
  vision}, pages 2425--2433.

\bibitem[{Bai et~al.(2023)Bai, Bai, Chu, Cui, Dang, Deng, Fan, Ge, Han, Huang
  et~al.}]{bai2023qwen}
Jinze Bai, Shuai Bai, Yunfei Chu, Zeyu Cui, Kai Dang, Xiaodong Deng, Yang Fan,
  Wenbin Ge, Yu~Han, Fei Huang, et~al. 2023.
\newblock Qwen technical report.
\newblock \emph{arXiv preprint arXiv:2309.16609}.

\bibitem[{Che et~al.(2024)Che, Wang, Liu, and
  Ma}]{che2024leveragingfoundationmodelsmultimodal}
Liwei Che, Jiaqi Wang, Xinyue Liu, and Fenglong Ma. 2024.
\newblock \href {https://arxiv.org/abs/2406.11048} {Leveraging foundation
  models for multi-modal federated learning with incomplete modality}.
\newblock \emph{Preprint}, arXiv:2406.11048.

\bibitem[{Che et~al.(2023)Che, Liu, Zhou, Ren, Zhou, Sheng, Dai, and
  Dou}]{che2023federated}
Tianshi Che, Ji~Liu, Yang Zhou, Jiaxiang Ren, Jiwen Zhou, Victor~S Sheng,
  Huaiyu Dai, and Dejing Dou. 2023.
\newblock Federated learning of large language models with parameter-efficient
  prompt tuning and adaptive optimization.
\newblock \emph{arXiv preprint arXiv:2310.15080}.

\bibitem[{Chen et~al.(2023)Chen, Zhang, Krompass, Gu, and
  Tresp}]{chen2023feddatapproachfoundationmodel}
Haokun Chen, Yao Zhang, Denis Krompass, Jindong Gu, and Volker Tresp. 2023.
\newblock \href {https://arxiv.org/abs/2308.12305} {Feddat: An approach for
  foundation model finetuning in multi-modal heterogeneous federated learning}.
\newblock \emph{Preprint}, arXiv:2308.12305.

\bibitem[{Chen and Zhang(2024)}]{chen2024disentanglement}
Jiayi Chen and Aidong Zhang. 2024.
\newblock On disentanglement of asymmetrical knowledge transfer for
  modality-task agnostic federated learning.
\newblock In \emph{Proceedings of the AAAI Conference on Artificial
  Intelligence}, volume~38, pages 11311--11319.

\bibitem[{Dai et~al.(2023)Dai, Li, Li, Tiong, Zhao, Wang, Li, Fung, and
  Hoi}]{dai2023instructblipgeneralpurposevisionlanguagemodels}
Wenliang Dai, Junnan Li, Dongxu Li, Anthony Meng~Huat Tiong, Junqi Zhao,
  Weisheng Wang, Boyang Li, Pascale Fung, and Steven Hoi. 2023.
\newblock \href {https://arxiv.org/abs/2305.06500} {Instructblip: Towards
  general-purpose vision-language models with instruction tuning}.
\newblock \emph{Preprint}, arXiv:2305.06500.

\bibitem[{Feng et~al.(2023)Feng, Bose, Zhang, Hebbar, Ramakrishna, Gupta,
  Zhang, Avestimehr, and
  Narayanan}]{feng2023fedmultimodalbenchmarkmultimodalfederated}
Tiantian Feng, Digbalay Bose, Tuo Zhang, Rajat Hebbar, Anil Ramakrishna, Rahul
  Gupta, Mi~Zhang, Salman Avestimehr, and Shrikanth Narayanan. 2023.
\newblock \href {https://arxiv.org/abs/2306.09486} {Fedmultimodal: A benchmark
  for multimodal federated learning}.
\newblock \emph{Preprint}, arXiv:2306.09486.

\bibitem[{Houlsby et~al.(2019)Houlsby, Giurgiu, Jastrzebski, Morrone,
  De~Laroussilhe, Gesmundo, Attariyan, and Gelly}]{houlsby2019parameter}
Neil Houlsby, Andrei Giurgiu, Stanislaw Jastrzebski, Bruna Morrone, Quentin
  De~Laroussilhe, Andrea Gesmundo, Mona Attariyan, and Sylvain Gelly. 2019.
\newblock Parameter-efficient transfer learning for nlp.
\newblock In \emph{International conference on machine learning}, pages
  2790--2799. PMLR.

\bibitem[{Hu et~al.(2021)Hu, Shen, Wallis, Allen-Zhu, Li, Wang, Wang, and
  Chen}]{hu2021lora}
Edward~J Hu, Yelong Shen, Phillip Wallis, Zeyuan Allen-Zhu, Yuanzhi Li, Shean
  Wang, Lu~Wang, and Weizhu Chen. 2021.
\newblock Lora: Low-rank adaptation of large language models.
\newblock \emph{arXiv preprint arXiv:2106.09685}.

\bibitem[{Kirkpatrick et~al.(2017)Kirkpatrick, Pascanu, Rabinowitz, Veness,
  Desjardins, Rusu, Milan, Quan, Ramalho, Grabska-Barwinska
  et~al.}]{kirkpatrick2017overcoming}
James Kirkpatrick, Razvan Pascanu, Neil Rabinowitz, Joel Veness, Guillaume
  Desjardins, Andrei~A Rusu, Kieran Milan, John Quan, Tiago Ramalho, Agnieszka
  Grabska-Barwinska, et~al. 2017.
\newblock Overcoming catastrophic forgetting in neural networks.
\newblock \emph{Proceedings of the national academy of sciences},
  114(13):3521--3526.

\bibitem[{Lai et~al.(2022)Lai, Dai, Singapuram, Liu, Zhu, Madhyastha, and
  Chowdhury}]{lai2022fedscale}
Fan Lai, Yinwei Dai, Sanjay Singapuram, Jiachen Liu, Xiangfeng Zhu, Harsha
  Madhyastha, and Mosharaf Chowdhury. 2022.
\newblock Fedscale: Benchmarking model and system performance of federated
  learning at scale.
\newblock In \emph{International conference on machine learning}, pages
  11814--11827. PMLR.

\bibitem[{Lester et~al.(2021)Lester, Al-Rfou, and Constant}]{lester2021power}
Brian Lester, Rami Al-Rfou, and Noah Constant. 2021.
\newblock The power of scale for parameter-efficient prompt tuning.
\newblock \emph{arXiv preprint arXiv:2104.08691}.

\bibitem[{Li et~al.(2023)Li, Li, Savarese, and Hoi}]{li2023blip}
Junnan Li, Dongxu Li, Silvio Savarese, and Steven Hoi. 2023.
\newblock Blip-2: Bootstrapping language-image pre-training with frozen image
  encoders and large language models.
\newblock In \emph{International conference on machine learning}, pages
  19730--19742. PMLR.

\bibitem[{Li et~al.(2020)Li, Sahu, Zaheer, Sanjabi, Talwalkar, and
  Smith}]{li2020federated}
Tian Li, Anit~Kumar Sahu, Manzil Zaheer, Maziar Sanjabi, Ameet Talwalkar, and
  Virginia Smith. 2020.
\newblock Federated optimization in heterogeneous networks.
\newblock \emph{Proceedings of Machine learning and systems}, 2:429--450.

\bibitem[{Liu et~al.(2024{\natexlab{a}})Liu, Li, Li, and Lee}]{liu2024improved}
Haotian Liu, Chunyuan Li, Yuheng Li, and Yong~Jae Lee. 2024{\natexlab{a}}.
\newblock Improved baselines with visual instruction tuning.
\newblock In \emph{Proceedings of the IEEE/CVF Conference on Computer Vision
  and Pattern Recognition}, pages 26296--26306.

\bibitem[{Liu et~al.(2024{\natexlab{b}})Liu, Li, Wu, and Lee}]{liu2024visual}
Haotian Liu, Chunyuan Li, Qingyang Wu, and Yong~Jae Lee. 2024{\natexlab{b}}.
\newblock Visual instruction tuning.
\newblock \emph{Advances in neural information processing systems}, 36.

\bibitem[{Lu et~al.(2022)Lu, Mishra, Xia, Qiu, Chang, Zhu, Tafjord, Clark, and
  Kalyan}]{lu2022learn}
Pan Lu, Swaroop Mishra, Tony Xia, Liang Qiu, Kai-Wei Chang, Song-Chun Zhu,
  Oyvind Tafjord, Peter Clark, and Ashwin Kalyan. 2022.
\newblock Learn to explain: Multimodal reasoning via thought chains for science
  question answering.
\newblock In \emph{The 36th Conference on Neural Information Processing Systems
  (NeurIPS)}.

\bibitem[{Lu et~al.(2021)Lu, Qiu, Chen, Xia, Zhao, Zhang, Yu, Liang, and
  Zhu}]{lu2021iconqa}
Pan Lu, Liang Qiu, Jiaqi Chen, Tony Xia, Yizhou Zhao, Wei Zhang, Zhou Yu,
  Xiaodan Liang, and Song-Chun Zhu. 2021.
\newblock Iconqa: A new benchmark for abstract diagram understanding and visual
  language reasoning.
\newblock In \emph{The 35th Conference on Neural Information Processing Systems
  (NeurIPS) Track on Datasets and Benchmarks}.

\bibitem[{Matena and Raffel(2022)}]{matena2022merging}
Michael~S Matena and Colin~A Raffel. 2022.
\newblock Merging models with fisher-weighted averaging.
\newblock \emph{Advances in Neural Information Processing Systems},
  35:17703--17716.

\bibitem[{McMahan et~al.(2017)McMahan, Moore, Ramage, Hampson, and
  y~Arcas}]{mcmahan2017communication}
Brendan McMahan, Eider Moore, Daniel Ramage, Seth Hampson, and Blaise~Aguera
  y~Arcas. 2017.
\newblock Communication-efficient learning of deep networks from decentralized
  data.
\newblock In \emph{Artificial intelligence and statistics}, pages 1273--1282.
  PMLR.

\bibitem[{Peng et~al.(2023{\natexlab{a}})Peng, Li, He, Galley, and
  Gao}]{peng2023instruction}
Baolin Peng, Chunyuan Li, Pengcheng He, Michel Galley, and Jianfeng Gao.
  2023{\natexlab{a}}.
\newblock Instruction tuning with gpt-4.
\newblock \emph{arXiv preprint arXiv:2304.03277}.

\bibitem[{Peng et~al.(2023{\natexlab{b}})Peng, Wang, Dong, Hao, Huang, Ma, and
  Wei}]{peng2023kosmos}
Zhiliang Peng, Wenhui Wang, Li~Dong, Yaru Hao, Shaohan Huang, Shuming Ma, and
  Furu Wei. 2023{\natexlab{b}}.
\newblock Kosmos-2: Grounding multimodal large language models to the world.
\newblock \emph{arXiv preprint arXiv:2306.14824}.

\bibitem[{Touvron et~al.(2023)Touvron, Lavril, Izacard, Martinet, Lachaux,
  Lacroix, Rozi{\`e}re, Goyal, Hambro, Azhar et~al.}]{touvron2023llama}
Hugo Touvron, Thibaut Lavril, Gautier Izacard, Xavier Martinet, Marie-Anne
  Lachaux, Timoth{\'e}e Lacroix, Baptiste Rozi{\`e}re, Naman Goyal, Eric
  Hambro, Faisal Azhar, et~al. 2023.
\newblock Llama: Open and efficient foundation language models.
\newblock \emph{arXiv preprint arXiv:2302.13971}.

\bibitem[{Wang et~al.(2024)Wang, Shen, He, Sun, Wang, Lyu, and
  Li}]{wang2024flora}
Ziyao Wang, Zheyu Shen, Yexiao He, Guoheng Sun, Hongyi Wang, Lingjuan Lyu, and
  Ang Li. 2024.
\newblock Flora: Federated fine-tuning large language models with heterogeneous
  low-rank adaptations.
\newblock \emph{arXiv preprint arXiv:2409.05976}.

\bibitem[{Wu et~al.(2023)Wu, Wang, Ge, Lu, Zhou, Shan, and Luo}]{wu2023pi}
Chengyue Wu, Teng Wang, Yixiao Ge, Zeyu Lu, Ruisong Zhou, Ying Shan, and Ping
  Luo. 2023.
\newblock $\pi$-tuning: transferring multimodal foundation models with optimal
  multi-task interpolation.
\newblock In \emph{Proceedings of the 40th International Conference on Machine
  Learning}, pages 37713--37727.

\bibitem[{Xu et~al.(2024)Xu, Shu, Mei, Xie, Fernando, Shou, and
  Tang}]{xu2024fedmllm}
Binqian Xu, Xiangbo Shu, Haiyang Mei, Guosen Xie, Basura Fernando, Mike~Zheng
  Shou, and Jinhui Tang. 2024.
\newblock Fedmllm: Federated fine-tuning mllm on multimodal heterogeneity data.
\newblock \emph{arXiv preprint arXiv:2411.14717}.

\bibitem[{Yang et~al.(2024)Yang, Long, Shen, Jiang, and
  Blumenstein}]{yang2024dual}
Yiyuan Yang, Guodong Long, Tao Shen, Jing Jiang, and Michael Blumenstein. 2024.
\newblock Dual-personalizing adapter for federated foundation models.
\newblock \emph{arXiv preprint arXiv:2403.19211}.

\bibitem[{Yi et~al.(2023)Yi, Yu, Wang, and Liu}]{yi2023fedlora}
Liping Yi, Han Yu, Gang Wang, and Xiaoguang Liu. 2023.
\newblock Fedlora: Model-heterogeneous personalized federated learning with
  lora tuning.
\newblock \emph{arXiv preprint arXiv:2310.13283}.

\bibitem[{Yin et~al.(2024)Yin, Fu, Zhao, Li, Sun, Xu, and Chen}]{yin2024survey}
Shukang Yin, Chaoyou Fu, Sirui Zhao, Ke~Li, Xing Sun, Tong Xu, and Enhong Chen.
  2024.
\newblock A survey on multimodal large language models.
\newblock \emph{National Science Review}, page nwae403.

\bibitem[{Yu et~al.(2023)Yu, Liu, Wang, Xu, and
  Liu}]{yu2023multimodalfederatedlearningcontrastive}
Qiying Yu, Yang Liu, Yimu Wang, Ke~Xu, and Jingjing Liu. 2023.
\newblock \href {https://arxiv.org/abs/2302.08888} {Multimodal federated
  learning via contrastive representation ensemble}.
\newblock \emph{Preprint}, arXiv:2302.08888.

\bibitem[{Zaken et~al.(2021)Zaken, Ravfogel, and Goldberg}]{zaken2021bitfit}
Elad~Ben Zaken, Shauli Ravfogel, and Yoav Goldberg. 2021.
\newblock Bitfit: Simple parameter-efficient fine-tuning for transformer-based
  masked language-models.
\newblock \emph{arXiv preprint arXiv:2106.10199}.

\bibitem[{Zhang et~al.(2024)Zhang, Vahidian, Kuo, Li, Zhang, Yu, Wang, and
  Chen}]{zhang2024towards}
Jianyi Zhang, Saeed Vahidian, Martin Kuo, Chunyuan Li, Ruiyi Zhang, Tong Yu,
  Guoyin Wang, and Yiran Chen. 2024.
\newblock Towards building the federatedgpt: Federated instruction tuning.
\newblock In \emph{ICASSP 2024-2024 IEEE International Conference on Acoustics,
  Speech and Signal Processing (ICASSP)}, pages 6915--6919. IEEE.

\bibitem[{Zhu et~al.(2023)Zhu, Chen, Shen, Li, and Elhoseiny}]{zhu2023minigpt}
Deyao Zhu, Jun Chen, Xiaoqian Shen, Xiang Li, and Mohamed Elhoseiny. 2023.
\newblock Minigpt-4: Enhancing vision-language understanding with advanced
  large language models.
\newblock \emph{arXiv preprint arXiv:2304.10592}.

\end{thebibliography}

\newpage
\clearpage
\appendix
\onecolumn
\appendix

\section{FedNano: Pseudocode Overview}
\label{app:code}
\begin{algorithm}
\caption{\textit{FedNano}. The $K$ clients are indexed by $k$; $R$ is the number of communication rounds, and $T$ is the number of local steps.}
\label{alg:fednano}
\textbf{Server Update:}
\begin{algorithmic}[1]
\State Randomly initialize $\mathcal{A}_I^0$ and $\mathcal{A}_T^0$ in \textit{NanoAdapter}, and distribute to clients
\For{$r = 1$ to $R$}
    \For{$k = 1$ to $K$ \textbf{in parallel}}
        \State $\theta_k^r \gets \text{ClientUpdate}(\theta^{r-1}_{global}, D_k)$
        \State Compute FIM $F_k$
    \EndFor
    \State $\theta^r_{global} \gets \text{ServerAgg}(\{\theta_k^r, F_k^r\})$ \Comment{
    Eq. \ref{eq:FIM}}
\EndFor
\end{algorithmic}
\textbf{Client Update}:
\begin{algorithmic}[1]
\State $\theta_k^{r-1} \gets \theta^{r-1}_{global}$
\For{local step $t = 1$ to $T$}
    \State Sample $\{(v_k,q_k,a_k)\}$ from $D_k$
    \State ${\theta_k^r}^{(t)} \gets \text{Optimization}({\theta_k^r}^{(t-1)}, v_k,q_k,a_k )$
\EndFor
\State \Return $\theta_k^r$
\end{algorithmic}
\end{algorithm}

\end{document}